\documentclass{bmvc2k}

\usepackage{multirow}
\usepackage{amssymb}

\let\oldciteauthor=\citeauthor
\def\citeauthor#1{\hypersetup{citecolor=black}\oldciteauthor{#1}}
\let\oldcite=\cite
\def\cite#1{\hypersetup{citecolor=red}\oldcite{#1}}

\title{Semi-supervised Feature-Level Attribute Manipulation for Fashion Image Retrieval}

\addauthor{Minchul Shin}{min.stellastra@gmail.com}{1}
\addauthor{Sanghyuk Park}{shine0624@gmail.com}{2}
\addauthor{Taeksoo Kim}{jazzsaxmafia@gmail.com}{2}

\addinstitution{
 Search Solutions Inc.\\
 Republic of Korea
}
\addinstitution{
 Naver Corp.\\
 Republic of Korea
}
\addinstitution{
 Naver Corp.\\
 Republic of Korea
}

\runninghead{Shin et al.}{Feature-Level Attribute Manipulation for Fashion}

\begin{document}

\maketitle

\begin{abstract}
With a growing demand for the search by image, many works have studied the task of fashion instance-level image retrieval (FIR). Furthermore, the recent works introduce a concept of fashion attribute manipulation (FAM) which manipulates a specific attribute (\emph{e.g} color) of a fashion item while maintaining the rest of the attributes (\emph{e.g} shape, and pattern). In this way, users can search not only ``the same'' items but also ``similar'' items with the desired attributes. FAM is a challenging task in that the attributes are hard to define, and the unique characteristics of a query are hard to be preserved. Although both FIR and FAM are important in real-life applications, most of the previous studies have focused on only one of these problem. In this study, we aim to achieve competitive performance on both FIR and FAM. To do so, we propose a novel method that converts a query into a representation with the desired attributes. We introduce a new idea of attribute manipulation at the feature level, by matching the distribution of manipulated features with real features. In this fashion, the attribute manipulation can be done independently from learning a representation from the image. By introducing the feature-level attribute manipulation, the previous methods for FIR can perform attribute manipulation without sacrificing their retrieval performance.
\end{abstract}

\section{Introduction}
\label{sec:intro}
The fashion instance-level image retrieval (FIR)~\cite{cross1,cross2,gajic2018cross,rec3} is a task to find a clothing item that is exactly ``the same'' as a given query. The FIR has been considered as a challenging task: the clothing items are highly deformable, the viewpoint severely differs, and minor differences are hard to catch. As an extension of FIR, the fashion attribute manipulation (FAM)~\cite{zhao2017memory,ak2018learning} aims to find ``similar'' clothing items by changing the unwanted attribute to the desired one, while preserving the rest of the attributes. This feature is particularly important in fashion search, because the users always desire to find clothing items with different attributes such as shape, color, or pattern (\textit{i.e., the user might want to search for a ``white checked blouse,'' not ``blue''}). To satisfy the user's needs in the fashion search, both FIR and FAM are important, and either one cannot be sacrificed.

From this perspective, the methods so far can be divided into two classes: \textit{FIR methods} and \textit{FAM methods}. The purpose of FIR methods is to find the same instances as in a given query. Most FIR methods use the ranking loss based on the instance labels for the training, sometimes exploiting the attribute labels as well. Since these methods were inherently designed in consideration of the FIR only, they are not capable of finding the item with desired characteristics by manipulating the attributes of the fashion item~\cite{ak2018learning}. Meanwhile, FAM methods aim to find similar-looking items by manipulating a specific attribute. Instead of creating pairs by instance labels, FAM methods create pairs by attribute labels inducing a model to focus on the attribute representation. However, creating pairs by attribute labels requires a dense annotation~\cite{zhao2017memory}, and having the same combinations of attributes does not always guarantee that the positive represents the same instance. For these reasons, FAM methods tend to sacrifice a pure instance retrieval performance. The simplest way we can think for solving both FIR, and FAM problem is to have both models for each specific tasks. However, in real-life applications dealing with billion-scale images, storing more than one features extracted from every single image is too expensive. Instead, we designed a feature-level attribute manipulation method that translates a learned feature directly into a manipulated feature representing desired attributes.

\noindent
\textbf{Main contributions.} In this paper, we propose a novel method that performs search by feature-level attribute manipulation (FLAM). The contribution is three fold: (1) We propose a new method for feature-level attribute manipulation, and we demonstrate its effectiveness by achieving competitive performance in both FIR and FAM tasks. (2) Unlike previous approaches, our method no longer requires a dense annotation of attributes for training. Instead, it transfers knowledge from learned attribute-specific embedders. (3) We show that a model for attribute manipulation can be trained in an semi-supervised manner, and a generative model can be used for image search with attribute manipulation.

\begin{figure}[t]
	\begin{center}
		\includegraphics[width=1.0\linewidth]{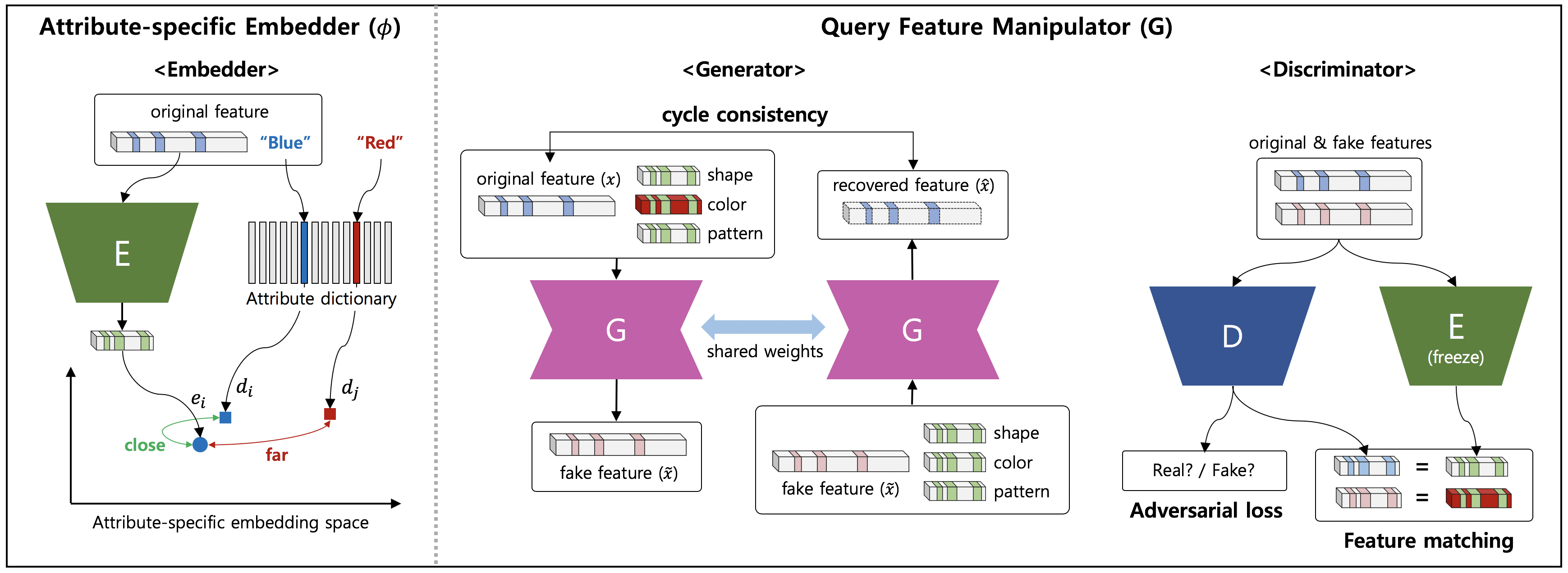}
	\end{center}
    \vspace{-7mm}
	\caption{The specific architecture of FLAM, which consists of two subnetworks: attribute-specific embedder, and query feature manipulator. The figure shows the scenario of manipulating color attribute while preserving the remaining attributes (shape and pattern). Note that the red-colored embedding vector is extracted from the randomly-sampled real feature that owns the desired color attribute (e.g., ``feature of red color item, if the query is blue ''). }
	\label{fig:main_diagram}
	\vspace{-5mm}
\end{figure}


\section{Related Work}
\label{sec:relatedworks}
The previous research on image retrieval in fashion domain can be categorized as follows: instance-based retrieval, attribute-based retrieval, and attribute manipulation.

\noindent
{\textbf{Instance-based retrieval.}}
The recent instance-level image retrieval methods have shown a dramatic increase in performance using advances in metric learning~\cite{gordo2017end,schroff2015facenet,sohn2016improved,yu2018hard,jun2019combination}. The metric learning focuses on the way it calculates loss by making a pair in an effective way. The instance ID is required for making a pair, and no additional labels are needed in general. In the fashion domain, several studies considered the cross-domain problem that aims to find in-shop images by querying street images in uncontrolled conditions~\cite{cross1,cross2,gajic2018cross,rec3}. \citeauthor{cross1}~\cite{cross1} learns a similarity measure between the street and in-shop domain, and \citeauthor{cross2}~\cite{cross2} uses human/clothing part alignment. \citeauthor{gajic2018cross}~\cite{gajic2018cross} trained a simple model with triplet loss and max pooling.

\noindent
{\textbf{Attribute-based retrieval.}}
The image retrieval in fashion domain is a distinct task in that fashion items have attributes. The attributes provide useful information to understand the fashion image in detail~\cite{att1,att2,att3}. For this reason, many publicly available fashion datasets provide the attribute labels as well~\cite{liu2016deepfashion,huang2015cross,ak2018efficient,corbiere2017leveraging}. We classify the methods that use attribute labels for training as attribute-based retrieval.
\citeauthor{liu2016deepfashion}~\cite{liu2016deepfashion} proposed FashionNet, which predicts landmarks and attributes simultaneously. A triplet loss is used for metric learning of pairwise clothes images. \citeauthor{huang2015cross}~\cite{huang2015cross} proposed a dual attribute-aware ranking network, which uses attribute-guided learning and triplet visual similarity constraint. \citeauthor{ret2}~\cite{ret2} defined a set of style-related attributes and trained an individual classifier for each style. The classifiers were then used for searching items by attributes. \citeauthor{attcnn1}~\cite{attcnn1} proposed a joint clothing detection model that combined a deep network with conditional random fields, considering the compatibility of clothing items and attributes. \citeauthor{attcnn5}~\cite{attcnn5} suggested an end-to-end training that localizes and ranks relative visual attributes in a weakly supervised manner. While visual attributes of fashion items provide high-level information, we found that the attribute labels in most of the fashion datasets are severely noisy, which may interfere with learning a good-quality image representation.

\noindent
{\textbf{Attribute manipulation.}}
Recently, a concept of interactive search in fashion domain has been introduced~\cite{kovashka2013attribute,modi2017confidence,plummer2019give}. The main idea is that users manipulate attributes of a query, and the search engine finds a fashion item with the desired attributes (\emph{e.g} green striped dress). Although attribute manipulation is particularly useful in fashion domain, we found that only a few papers have addressed this issue~\cite{zhao2017memory,ak2018learning}. \citeauthor{zhao2017memory}~\cite{zhao2017memory} proposed to use a memory block to obtain a manipulated feature. The feature was combined with an attribute-specific representation from the memory block. Triplet pairs were constructed based on a combination of attribute labels, which requires a dense annotation for attributes. \citeauthor{ak2018learning}~\cite{ak2018learning} suggested a region-aware attribute manipulation by the integration of the attribute activation maps. The extracted features per regions are then combined to construct the global representation.

While aforementioned methods achieved competitive results on their tasks, more fundamental problems has not been solved. First, despite the powerful performance of the FIR methods (instance-based, and attribute-based), the methods were not designed to search for items with desired attributes. Therefore, the FIR methods have no way to find users similarly-looking items with the desired attributes, which is a huge disadvantage particularly in fashion domain. Second, the methods for attribute manipulation enable search by changing the attributes, but the image representation of these methods is not specialized for the FIR task. Therefore the FAM methods show worse performance in finding the same instance items. To remedy this, we propose a method that can perform attribute manipulation without sacrificing the retrieval performance by manipulating queries at the feature level.

\begin{figure}[t]
	\begin{center}
		\includegraphics[width=1.0\linewidth]{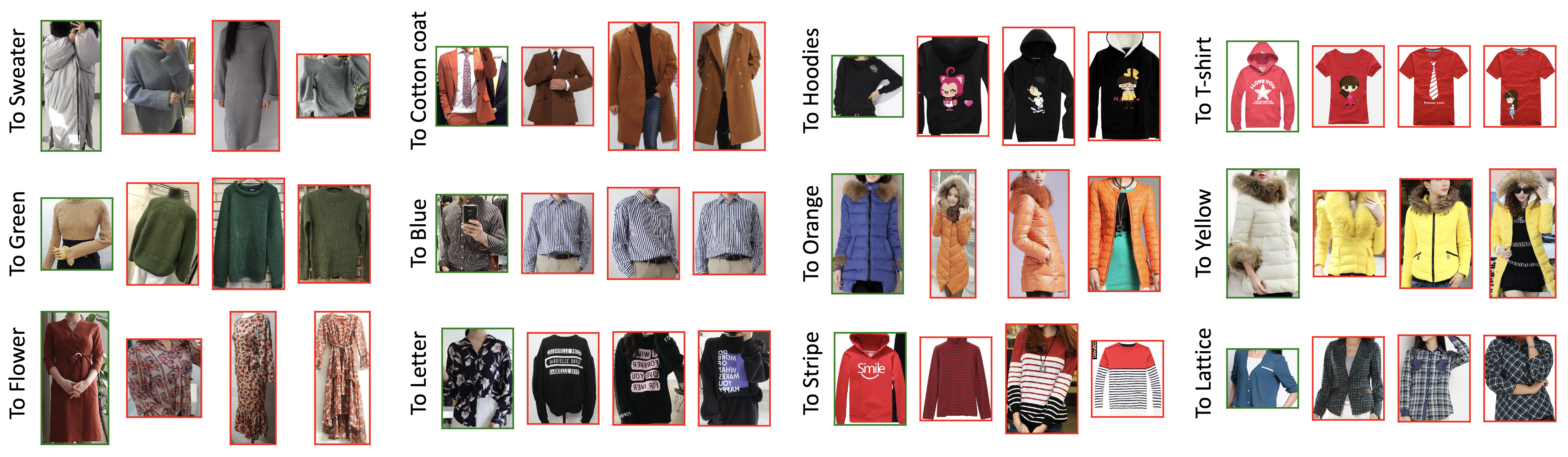}
	\end{center}
    \vspace{-7mm}
	\caption{Top-3 retrieval results after the query attribute manipulation. The green-bordered images are the query, and the red-bordered images are retrieved images from the gallery. The first and second columns are the results on NFD, and the third and fourth columns are the results on DARN. From top to bottom, each row targets to manipulate shape, color, and pattern attribute, respectively.}
	\label{fig:top_k_visualization}
	\vspace{-5mm}
\end{figure}

\section{Feature-level Attribute Manipulation (FLAM)}
\label{sec:method}
Our objective is to train a network that converts the feature of the given query into the feature with the desired attributes. We refer to it as \textit{feature-level attribute manipulation}, because the attribute of a learned feature is manipulated directly at the feature level, and no representation learning from the image is needed. The feature-level attribute manipulation has several advantages. First, since the features learned by FIR methods are used as the input for the manipulation, the quality of the feature performing FIR can be perfectly preserved. Second, the learned features include high-level information comparing to the images in RGB space, and the network can be trained with a smaller number of data, even with the sparsely labeled attribute information in the fashion datasets. For manipulating the attributes at the feature level, our main idea is to synthesize the manipulated feature with a generative adversarial network~\cite{goodfellow2014generative}, by simply matching the distribution of the manipulated feature with the distribution of the real data. We split the training into two stages as described in Figure~\ref{fig:main_diagram}: \textit{(1) embedding stage} and \textit{(2) manipulating stage}. We explain the training procedure of each stage in detail, followed by the way of synthesizing the manipulated feature that represents the desired attributes.

\subsection{Attribute-specific Embedder}
At the embedding stage, we aim to embed a learned feature to an attribute-specific embedding space defined as ${e\in R^{k}}$, where $k$ is the dimension of the embedding space~\cite{vasileva2018learning}. We assume that the attribute-specific information (such as \textit{shape, color, and pattern}) is successfully encoded in a learned feature when it is learned by FIR methods. For a learned feature $x$, we find the embedding vectors of target attribute $a$ as $e_{a} = \phi_{a}(x;\theta)$. The goal is to learn parameter $\theta$ of mapping $\phi$ such that embedding vectors $e_a$ and $e^{+}_a$ extracted from samples of the same attribute class $i$ are projected into as close location as possible in the embedding space and in far distance from $e^{-}_a$ of a different attribute class. We use the embedding layer, which stores the fixed size of dictionary vector $d_{a}$. The general form of the triplet loss is formulated as \eqref{eq:eq1}

\begin{equation}
    \label{eq:eq1}
    \ell(f,f^+,f^-)=max\{0, dist(f,f^+)-dist(f,f^-)+\mu\},
\end{equation}
where $\mu$ is a margin, $dist$ is a cosine similarity distance. Finally, we use a triplet loss using both $e_{a}$ and $d_{a}$ as anchors for training $\phi$. The final loss is formulated as \eqref{eq:eq2},
\begin{equation}
    \label{eq:eq2}
    \mathcal{L}_{emb}= \ell(d_{a},e^{+}_{a},e^{-}_{a}) + \ell(e_{a},d^{+}_{a},d^{-}_{a}),
\end{equation}
where $^-$ symbol means that the vector is extracted from the sample of a different attribute class compared with the anchor. Multiple attribute-specific embedders $\phi$ are trained are trained for every single attribute type, and they work as a teacher of a discriminator by transferring knowledge in the following stage.

\subsection{Query Feature Manipulator}
At the manipulating stage, our main interest is to synthesize a manipulated feature $\Tilde{x}$ which follows the distribution of real features $x$. To avoid ambiguity, we define the target attribute type to manipulate as $a$ (\emph{e.g} color), and the remaining attribute types to preserve as $\bar{a}_{1},...,\bar{a}_{n}$ (\emph{e.g} shape and pattern). The embedding vectors are defined as $e_{a}$ and $e_{\bar{a}_1},...,e_{\bar{a}_n}$  the target and remaining attribute types, respectively. Given a pair of learned features ($x$, $x^{-}$) sampled from different attribute classes of $a$, our goal is to train a generator $G(x, \phi_{a}(x^{-}))\rightarrow\Tilde{x}$ that converts $x$ into an attribute-manipulated feature $\Tilde{x}$ that represents the target attribute of $^{-}$ while preserving the remaining attributes of $x$.

\noindent
\textbf{Adversarial loss.} To make $\Tilde{x}$ synthesized by $G$ follow the distribution of real features, we adopt an adversarial loss~\cite{goodfellow2014generative} as defined
\begin{equation}
    \label{eq:eq_adv_loss}
    \mathcal{L}_{adv}=\mathbb{E}_{x}[\log D(x)] + \mathbb{E}_{x,x^{-}}[\log (1-G(x,e^{-}_a))],
\end{equation}
where $G$ synthesizes $\Tilde{x}=G(x,e^{-}_a)$ conditioned on query feature $x$ and attribute-specific embedding vector $e^{-}_a=\phi_{a}(x^{-})$. We concatenate $x$ and $e^{-}_a$ to feed $G$. The adversarial loss forces the generator $G$ to mimic the real data distribution by playing a minimax game.

\noindent
\textbf{Feature matching loss.} We aim $\Tilde{x}$ to have desired attribute class of $a$, not affecting the remaining attributes $\bar{a}_{1},...,\bar{a}_{n}$. Our strategy is to transfer the attribute-specific knowledge learned by $\phi$ by matching outputs of discriminator $D$ with the embedding vectors~\cite{salimans2016improved}. To achieve this condition, the last layer of $D$ is split into two branches, and each branch produces 1-dimensional vector to discriminate between real and fake, and $(n \times k)$-dimensional vector for feature matching with $e$, where $e\in R^k$. Letting $f(x)$ denote the outputs of $D$ for feature matching, for given attribute-specific embedders $\phi_{a}$ and $\phi_{\bar{a_{1}},...,\bar{a_{n}}}$, our objective is defined as
\begin{equation}
    \label{eq:eq_match_loss}
    \begin{split}
    \mathcal{L}_{match} = \mathbb{E}_{x,\Tilde{x}}[\lVert f(x) - [\{\phi_{a}(x)\}_{target} \oplus \{\phi_{\bar{a_1}}(x) \oplus,..., \phi_{\bar{a_n}}(x)\}_{remain}] \lVert]_{real} \\
    + \mathbb{E}_{x,x^{-},\Tilde{x}}[\lVert f(\Tilde{x}) - [\{\phi_{a}(x^{-})\}_{target} \oplus \{\phi_{\bar{a_1}}(x) \oplus,..., \phi_{\bar{a_n}}(x)\}_{remain}] \lVert]_{fake},
    \end{split}
\end{equation}
where $\oplus$ is a concatenation operation. The feature matching term for the remaining attributes can be omitted, but we found it harmful for the performance, as shown in Table~\ref{tab:am_performance}.

\noindent
\textbf{Cycle consistency loss.} Recently, a cycle consistency has proved its effectiveness in preserving the content of the original input~\cite{kim2017learning,zhu2017unpaired}. To preserve unique characteristics of a query item, we apply the cycle-consistency loss to the generator, defined as
\begin{equation}
    \label{eq:eq_cycle_loss}
    \mathcal{L}_{cycle} = \mathbb{E}_{x,x^{-}}[\lVert x - G(G(x,e^{-}),e) \lVert]
\end{equation}
An interesting point is that a cosine similarity between original feature $x$ and recovered feature $\hat{x}=G(G(x,e^{-}),e)$ can be used for an approximate measure of convergence~\cite{berthelot2017began}. We found a consistent tendency that a model showing a higher cosine similarity between $x$ and $\hat{x}$ achieves a better top-k accuracy in the attribute manipulation task.

\subsection{Query Manipulation}
To search with a query after the attribute manipulation, a manipulated query is obtained by $G(x,d_{a_i})\rightarrow\Tilde{x}$, where $d$ is an attribute-specific dictionary vector of $i$-th class of attribute type $a$ which has been learned at embedding stage. The attribute-specific dictionary vector $d_{a_i}$ can be considered as a centroid of a feature cluster of $i$-th attribute class. Finally, a query having the desired attributes can be synthesized by choosing $d$ of a desired attribute class $i$.

\vspace{-4mm}
\section{Experiments}
\label{sec:experiment}
In this section, we describe the datasets, evaluation metrics, and implementation details followed by quantitative and qualitative results. In this study, we target three most representative fashion attribute types, $a \in \{shape, color, pattern\}$, for attribute manipulation. When we target a specific attribute, we call it the target attribute (\emph{e.g} color), and the others as the remaining attributes (\emph{e.g} shape and pattern), for clarity.

\noindent
\textbf{Datasets.} Although many fashion-related datasets have been released, we found few fashion datasets provide reliable attribute labels. We used DARN and Shopping100k to evaluate and train the attribute-specific embedder. The top-k classes in each attribute type, sorted by the number of images included, were used to avoid the long-tail distribution. The detailed classes and labels of the datasets are described in Table~\ref{tab:attr_define}. To evaluate FAM, we randomly sampled 5K and 2K images from DARN and Shopping100k, respectively. We only picked the images labeled with all three attributes densely. The remaining images were used for the training, because FLAM does not require a dense annotation. To manipulate the training query features, we used private datasets, Naver Fashion Dataset (NFD)~\cite{navershopping}. The NFD is a large-scale fashion dataset collected from the Naver shopping pages, which includes more than 390K instances of 4-million high-resolution images. The instance pair label was annotated by the human experts to ensure its reliability. While NFD provides no attribute labels, we could use the entire images of the NFD by using the attribute-specific embedder as a pseudo-labeler by means of feature matching.

\begin{table}[t]
\footnotesize
\centering
\begin{tabular}{|c|c|}
\hline
Attribute types & Top-10 classes (DARN) \\ \hline \hline
Shape & sweater, hoodies, ... , cotton\_clothes (clothes\_category) \\
Color & black, red, ... , rose\_red (clothes\_color) \\
Pattern & solid\_color, flower, ... , abstract (clothes\_pattern) \\
\hline \hline
Attribute types & Top-10 classes (Shopping100k) \\ \hline \hline
Shape & T-shirt, Shirt, ... , Tracksuit\_Bottoms (category) \\
Color & Black, Navy, ... , Olive (color) \\
Pattern & Plain, Print, ... , Colourful (pattern) \\
\hline
\end{tabular}
\caption{Top-10 attribute classes used for experiment. The label in bracket is the original label in the datasets.}
\label{tab:attr_define}
\vspace{-5mm}
\end{table}


\noindent
\textbf{Evaluation metrics.} For the quantitative experiments, we define the evaluation metrics for both FIR and FAM as follows. For FIR task, recall(R@k)~\cite{gordo2017end} was measured. For FAM task, top-k accuracy(T@k)~\cite{huang2015cross} was measured. To measure top-k accuracy, the search system retrieves top-k matches after performing the attribute manipulation on a given query. If a match is found that has the same attribute combination (shape, color, and pattern) as the manipulated query, we count it as a hit, otherwise a miss.

\vspace{-3mm}
\subsection{Implementation Details}
When training an attribute-specific embedder, we set the dimension $k$ of the embedding space to 32. By squeezing the attribute-related representation in fewer dimensions, we prevent the embedding vector from including information irrelevant to the target attribute. When training a query feature manipulator, $G$, a randomly chosen embedding vector $e^{-}$ is given to synthesize a manipulated feature. However, sometimes it becomes a problem when attributes of a fashion item are highly entangled. For example, if a query item is a brown leather jacket and the embedding vector represents ``pink'', there barely exists such cloth in the real world. In this case, a manipulator is confused, because it has never seen such data in the real distribution. To avoid this, we used a technique called \textit{online sampling(OS)}. In detail, for a given batch, $x^{-}$ is sampled to have as close distance in the embedding space for $\bar{a_1},...,\bar{a_n}$ as possible. By choosing $x^-$ among the samples having the same remaining attributes as $x$ as possible, we could avoid forcing the manipulator to synthesize unrealistic features. Note that a pair $(x,x^{-})$ is sampled to have different target attribute classes for $a$. The effect of online sampling is shown in Table~\ref{tab:am_performance}.

\subsection{Quantitative Results} In this section, we report the performance evaluation on both FIR and FAM, comparing them to the baseline methods. The evaluation is performed using the public datasets: DeepFashion, DARN, and Shopping100k.

\begin{table}[b]
\vspace{-5mm}
\footnotesize
\centering
\begin{tabular}{|c|c|c|c|c|c|c|c|c|}
\hline
\multirow{2}{*}{Method} & \multicolumn{4}{c|}{DeepFashion(In-shop)} & \multicolumn{4}{c|}{DeepFashion(C2S)} \\ \cline{2-9}
                        & R@1     & R@5     & R@20    & R@50    & R@1     & R@5     & R@20    & R@50     \\ \hline \hline
WTBI~\cite{chen2012describing}                    & 0.347 & 0.424 & 0.506 & 0.541 & 0.024 & 0.035 & 0.063 & 0.087  \\
DARN~\cite{huang2015cross}                    & 0.381 & 0.547 & 0.675 & 0.716 & 0.036 & 0.063 & 0.111 & 0.152  \\
FashionNet~\cite{liu2016deepfashion}              & 0.533 & 0.673 & 0.764 & 0.793 & 0.073 & 0.121 & 0.188 & 0.226  \\
FashionNet+Poselets~\cite{yang2011articulated}     & 0.424 & 0.600 & 0.699 & 0.736 & 0.043 & 0.079 & 0.139 & 0.182  \\
FashionNet+Joints~\cite{yang2011articulated}       & 0.414 & 0.605 & 0.682 & 0.723 & 0.044 & 0.075 & 0.136 & 0.180  \\ \hline
Ours(ResNet50-SPoC)     & \textbf{0.887} & \textbf{0.961} & \textbf{0.984} & \textbf{0.991} & \textbf{0.265} & \textbf{0.497} & \textbf{0.664} & \textbf{0.755} \\
\hline
\end{tabular}
\caption{Instance-level image retrieval performance (R@k) on DeepFashion dataset. C2S indicates the consumer-to-shop subset of DeepFashion. }
\label{tab:ir_performance}
\end{table}

\noindent
\textbf{Baselines.}
Because the previous works on attribute manipulation required a dense annotation, and even the target attributes were different, it is unfair to compare the methods directly. Furthermore, no paper so far has discussed feature-level attribute manipulation. Instead, we propose the following baselines to evaluate our method. (1) \textit{Cls-Concat}: a concatenation of features from three classifiers for shape, color, and pattern. The network has two fully connected layers followed by a ReLU and a dropout~\cite{srivastava2014dropout}, and trained by cross-entropy loss. We fixed the size of the intermediate layer to 341. (2) \textit{Emb-Concat}: a concatenation of features from three attribute-specific embedders for shape, color, and pattern. The triplet loss is used, and the size of the embedding vectors is fixed to 341. The final dimension after the concatenation in both baselines is 1023 ($341\times3$), which is similar to FLAM. We consider these baselines as an upper bound in performance for FAM.

\noindent
\textbf{FIR performance.} Because the FLAM performs the attribute manipulation at a feature level, learning a good representation is important. We used cross-entropy loss and triplet loss together using ResNet50~\cite{he2016deep} as a backbone to train the feature extractor. We sampled thousand instances sorted by the number of images included in instance id to apply the cross-entropy loss, and then entire images were used when back-propagating with triplet loss. The feature maps are pooled using the sum-pooled convolutional features (SPoC)~\cite{babenko2015aggregating}. Note that the FIR performance can be improved by a careful aggregation of the pooling methods~\cite{jun2019combination}, although we omitted the experiments on it. By training with the instance labels only, we could obtain competitive recall on DeepFashion dataset in Table~\ref{tab:ir_performance}. The FLAM can be seen as a postprocessing method of a learned feature for the attribute manipulation. We consider the FIR result of our feature extractor as the performance of the FLAM on FIR.

\begin{table}[t]
\footnotesize
\centering
\begin{tabular}{|c|c|c|c|c|c|c|c|c|c|}
\hline
\multirow{2}{*}{Method} & \multirow{2}{*}{Dim} & \multicolumn{4}{c|}{DARN~\cite{huang2015cross} (T@50)}       & \multicolumn{4}{c|}{Shopping100k~\cite{ak2018efficient} (T@50)} \\ \cline{3-10}
                                        &                      & S & C & P & All   & S  & C  & P & All   \\ \hline \hline
AMNet*~\cite{zhao2017memory}            & n/a                  & -     & -     & -     & 0.668 & -     & -     & -     & 0.711 \\
FashionSearchNet*~\cite{ak2018learning} & n/a                  & -     & -     & -     & -     & -     & -     & -     & 0.732 \\ \hline
Cls-Concat                              & 1023                 & \textbf{0.896} & \textbf{0.947} & 0.698 & \textbf{0.898} & 0.886 & 0.895 & 0.802 & 0.874  \\
Emb-Concat                              & 1023                 & 0.758 & 0.769 & \textbf{0.888} & 0.779 & \textbf{0.916} & \textbf{0.925} & \textbf{0.893} & \textbf{0.915}  \\ \hline
FLAM(S/-/Adv)                           & 1024                 & 0.472 & 0.560 & 0.375 & 0.505 & 0.517 & 0.338 & 0.430 & 0.424 \\
FLAM(M/-/-)                             & 1024                 & 0.612 & 0.874 & 0.500 & 0.730 & 0.602 & 0.587 & 0.483 & 0.573  \\
FLAM(M/-/Adv)                           & 1024                 & 0.799 & 0.872 & 0.521 & 0.802 & 0.640 & 0.749 & 0.628 & 0.685  \\
FLAM(M/OS/Adv)                          & 1024                 & 0.824 & 0.867 & 0.624 & 0.822 & 0.633 & 0.768 & 0.633 & 0.691  \\ \hline
\end{tabular}
\caption{Top-50 accuracy for FAM. The notation indicates M/S: feature matching with/without remaining attributes, Adv: adversarial loss, and OS: online sampling. S, C, and P stand for Shape, Color and Pattern. * means that the value should be used for reference only, because the experiments for these methods used different combinations of attributes. }
\label{tab:am_performance}
\vspace{-4mm}
\end{table}

\begin{table}[t]
\footnotesize
\centering
\begin{tabular}{|c|c|c|c|c|c|c|c|c|c|}
\hline
\multirow{2}{*}{Method} & \multirow{2}{*}{Dim} & \multicolumn{4}{c|}{FIR(DeepFashion(C2S)~\cite{liu2016deepfashion})} & \multicolumn{4}{c|}{FAM(DARN~\cite{huang2015cross})} \\ \cline{3-10}
 &  & R@1 & R@5 & R@20 & R@100 & T@1 & T@5 & T@20 & T@100 \\ \hline \hline
Cls-Concat & 1023 & 0.043 & 0.092 & 0.164 & 0.301 & 0.284 & \textbf{0.660} & \textbf{0.838} & \textbf{0.932} \\
Emb-Concat & 1023 & 0.020 & 0.043 & 0.083 & 0.180 & \textbf{0.340} & 0.632 & 0.732 & 0.812 \\ \hline
\multicolumn{1}{|l|}{FLAM(M/OS/Adv)} & \multicolumn{1}{l|}{1024} & \multicolumn{1}{l|}{ \textbf{0.095} } & \multicolumn{1}{l|}{ \textbf{0.200} } & \multicolumn{1}{l|}{ \textbf{0.329} } & \multicolumn{1}{l|}{ \textbf{0.519} } & \multicolumn{1}{l|}{ 0.176 } & \multicolumn{1}{l|}{ 0.458 } & \multicolumn{1}{l|}{ 0.707 } & \multicolumn{1}{l|}{ 0.884 } \\
\hline
\end{tabular}
\caption{Simultaneous comparison between FIR and FAM performances.}
\label{tab:ir_am_compare}
\vspace{-4mm}
\end{table}

\begin{table}[!t]
\footnotesize
\centering
\begin{tabular}{|c|c|c|}
\hline
Backbone          & R@20 & T@20 \\ \hline \hline
ResNet50~\cite{he2016deep}          & 0.235 & 0.733    \\
SE-ResNet50~\cite{hu2018squeeze}       & 0.205 & 0.744    \\
SE-ResNeXt(32x4d)~\cite{hu2018squeeze,xie2017aggregated} & 0.240 & \textbf{0.780}    \\
InceptionResNetV2~\cite{szegedy2017inception} &  0.218   & 0.719    \\
DenseNet-121~\cite{huang2017densely}      & \textbf{0.255} & 0.691    \\
\hline
\end{tabular}
\quad
\begin{tabular}{|c|c|c|c|c|}
\hline
\multicolumn{5}{|c|}{DARN~\cite{huang2015cross} (Classification accuracy)}\\ \hline
query       &   S   &   C   &   P   & Avg.diff  \\ \hline \hline
original    & 0.352 & 0.730 & \textbf{0.768} & -        \\
manipulated & \textbf{0.577} & \textbf{0.866} & 0.714 & \textcolor{green}{+0.102} \\ \hline \hline
\multicolumn{5}{|c|}{Shopping100k~\cite{ak2018efficient} (Classification accuracy)}   \\ \hline
query       &  S  &  C  &  P  &  Avg.diff  \\ \hline \hline
original    & 0.819 & 0.675 & \textbf{0.814} & -    \\
manipulated & \textbf{0.843} & \textbf{0.941} & 0.749 & \textcolor{green}{+0.075} \\
\hline
\end{tabular}
\caption{R@k and T@k on different backbones (k=20) and classification accuracy after the attribute manipulation. The evaluation dataset is the same as Table~\ref{tab:ir_am_compare}.}
\label{tab:backbone_accdrop}
\vspace{-2mm}
\end{table}

\noindent
\textbf{FAM performance.} The Table~\ref{tab:am_performance} reports the top-50 accuracy of baselines and our method. Compared to the baselines, FLAM shows comparable performance on both datasets, despite that the query feature manipulator of FLAM was trained in a semi-supervised manner. We also achieved better results when auxiliary embedders of the remaining attributes and the adversarial loss were both used. Without the adversarial loss, the cosine similarity between the original feature and the recovered feature is low, which indicates that the manipulator suffers from recovering distribution of the original data. The use of online sampling technique consistently improved the performance. In Table~\ref{tab:ir_am_compare}, we compare FIR and FAM simultaneously, as our purpose is to obtain high performance in both tasks. As we expected, our method shows outperforming result in FIR task, as the learned feature was trained for it.

\noindent
\textbf{Quality of features, and robustness to backbones.} The quality of the manipulated features and the robustness depending on different backbones were investigated in Table~\ref{tab:backbone_accdrop}. We trained a simple classifier, and measured the classification accuracy on the original features. Then, the query was manipulated, and the accuracy on manipulated feature was measured. The result shows that the accuracy increases after the attribute manipulation, which means that the query attribute was well-manipulated as we intended. Moreover, FLAM shows consistent results on both FIR and FAM, regardless of the backbone, which shows the robustness of our method in the network structure.

\begin{figure}[t]
	\begin{center}
		\includegraphics[width=0.95\linewidth]{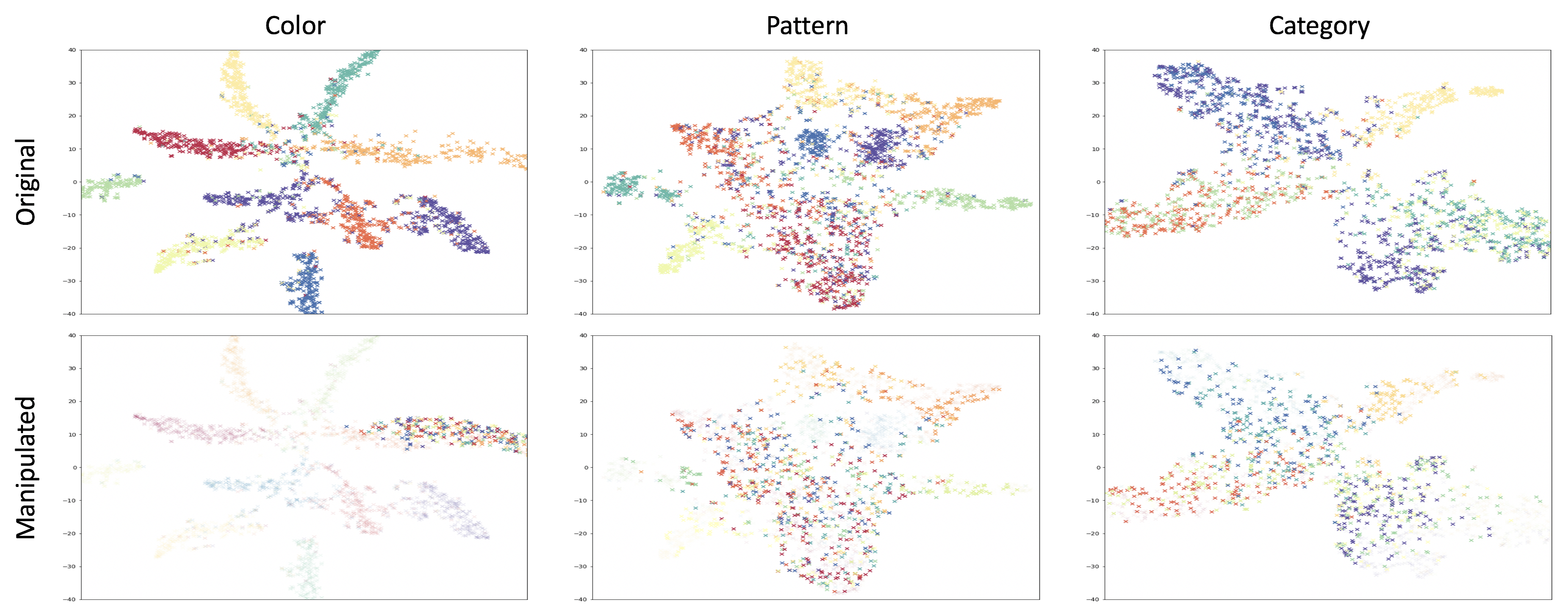}
	\end{center}
    \vspace{-7mm}
	\caption{t-SNE visualization of the attribute-specific embedding vectors on the embedding space. Among 3K samples of the original feature, the color attribute of randomly sampled 1K features is manipulated to represent a specific color. The color of the mark represents the original attribute classes on attribute types.}
	\label{fig:tsne}
	\vspace{-5mm}
\end{figure}

\subsection{Qualitative Results}
\vspace{-1mm}
In this section, we visualize the top-k ranking results after the attribute manipulation in Figure~\ref{fig:top_k_visualization}, and t-SNE~\cite{maaten2008visualizing} visualization of attribute-specific embedding vectors in Figure~\ref{fig:tsne}.
\newline
\textbf{Top-k ranking result.} The Figure~\ref{fig:top_k_visualization} shows the top-3 ranking results after the query attribute manipulation. The results show that FLAM retrieves the items having the desired attribute as we intended, while maintaining the unique characteristics of the query that are hard to define with shape, color, and pattern only. For example, if the query is a purple-colored slim-fit padding jumper with fur, and we manipulate color, the retrieved items maintain the attributes ``slim-fit'' and ``fur'', which have not been learned. We report top-3 ranking results, because we found that the evaluation metric of top-k accuracy on shape, color, and pattern have a limitation in reflecting the aforementioned aspect.
\newline
\textbf{t-SNE visualization.} To analyze our method in the feature space, we visualize the distribution of the attribute-specific embedding vectors of shape, color, and pattern attributes using t-SNE. We manipulate the query features to have the same color, while maintaining the remaining attributes (shape and pattern). After manipulating the color attribute, the manipulated feature is fed to the attribute-specific embedder once again to obtain the embedding vector of the manipulated feature. The Figure~\ref{fig:tsne} shows that the color distribution of the manipulated features is focused on a certain location, which indicates that the color is successfully manipulated to represent a certain color. The distribution of shape and pattern attributes shows a similar distribution with one of the original features, which means that the remaining attributes are preserved while manipulating the color.

\vspace{-4mm}
\section{Conclusions}
\label{sec:conclusions}
\vspace{-1mm}
In this paper, we have addressed the problem of attribute manipulation at the feature level. Unlike previous approaches, feature-level attribute manipulation (FLAM) translates the attributes from a learned feature itself, which makes the attribute manipulation process completely independent from image-based representation learning. We demonstrated competitive results on both FIR and FAM tasks with FLAM, and showed its robustness in application of different backbones. We showed that the generative model trained in a semi-supervised manner can be used for attribute manipulation, by simply matching the distribution of the manipulated features with the real features.

\newpage
\bibliography{paper}
\end{document}